# A neuromorphic hardware architecture using the Neural Engineering Framework for pattern recognition

Runchun Wang, Chetan Singh Thakur, Tara Julia Hamilton, Jonathan Tapson, André van Schaik
The MARCS Institute, University of Western Sydney, Sydney, NSW, Australia
mark.wang@uws.edu.au

*Abstract*—We present a hardware architecture that uses the Neural Engineering Framework (NEF) to implement large-scale neural networks on Field Programmable Gate Arrays (FPGAs) for performing pattern recognition in real time. NEF is a framework that is capable of synthesising large-scale cognitive systems from subnetworks. We will first present the architecture of the proposed neural network implemented using fixed-point numbers and demonstrate a routine that computes the decoding weights by using the online pseudoinverse update method (OPIUM) in a parallel and distributed manner. The proposed system is efficiently implemented on a compact digital neural core. This neural core consists of 64 neurons that are instantiated by a single physical neuron using a time-multiplexing approach. As a proof of concept, we combined 128 identical neural cores together to build a handwritten digit recognition system using the MNIST database and achieved a recognition rate of 96.55%. The system is implemented on a state-of-the-art FPGA and can process 5.12 million digits per second. The architecture is not limited to handwriting recognition, but is generally applicable as an extremely fast pattern recognition processor for various kinds of patterns such as speech and images.

## 1. Introduction

Neural networks have been proved to be powerful tools for real world tasks, such as pattern recognition, classification, regression, and prediction. However, their high computational demands are not ideally suited to modern computer architectures. This constraint has so far often prohibited their use in applications that need real-time control, such as interactive robotic systems. On the other hand, scientists have been developing hardware platforms that are optimised for neural networks over the past two decades (Vogelstein et al., 2007; Boahen, 2006; Pfeil et al., 2013; Wang et al., 2014d). However, these systems are not capable of synthesising large-scale neural networks for these real world tasks from subnetworks and therefore are not very suitable, as pointed out by Tapson *et al*. (Tapson et al., 2013).

Here, we present a generic hardware architecture that uses the Neural Engineering Framework (NEF) (Eliasmith and Anderson, 2003) to implement large-scale neural networks on FPGAs, which are capable of processing up to millions of pattern recognitions in real time. The NEF, which was first introduced in 2003, is a framework that is capable of building large systems from subnetworks with a standard three-layer neural structure (the first layer contains the input neurons; the second layer is a hidden layer, which consists of a large number of non-linear neurons; and the third layer is the output layer, which consists of linear neurons). The NEF has been used to construct SPAUN, which is the first brain model, implemented in software and is capable of performing cognitive tasks (Eliasmith et al., 2012). This demonstrates that the NEF is a powerful tool for synthesising large-scale cognitive systems.

We have previously presented a compact neural core architecture specifically for FPGA implementation of large NEF networks (Wang et al., 2014a). In this paper, we present an application that uses this neural core to build pattern recognition systems. The outline for this paper is as follows: Section 2.1 introduces the basic concepts of the NEF; the algorithm and theory is presented in Section 2.2; the hardware implementation is presented in Section 2.3; the performance for different design choices will be thoroughly compared in Section 3; in section 4 we compare our work with other solutions and discuss future works.

## 2. Materials and methods

### 2.1 Background

In this section, we review the theoretical framework of a typical NEF system, which encodes an input stimulus into a spiking rate of neurons of a heterogeneous population and decodes the desired function by linearly combining the responses of these neurons. The topology of the NEF network is illustrated in Figure 1. A NEF network performs three tasks to calculate a desired function $f(X)$:

**1. Encoding**: An encoder will have a fixed random weight (RW) for each hidden layer neuron, and multiplies the input stimulus by this weight. The firing rate of individual neurons is a nonlinear function of the input stimulus weighted by the random weights. The parameters of the neurons are also randomised, so that each neuron in the hidden layer exhibits a distinct tuning curve. An example of such tuning curves is shown in Figure 2.

**2. Decoding:** The activity, $H$, of the hidden neurons (i.e. the spike rate of each neuron) can be measured over the desired range of input values $X$. The output of each neuron will be multiplied by their decoding weights such that $WH = f(X) = Y$. Since this is a linear system, these weights can be found by calculating $W = YH^+$, where $H^+$ is the Moore-Penrose pseudo-inverse (Penrose and Todd, 1955) of $H$.

**3. Averaging:** The output of the system, $Y(t)$, is the linear sum of the weighted spike trains from the neurons.

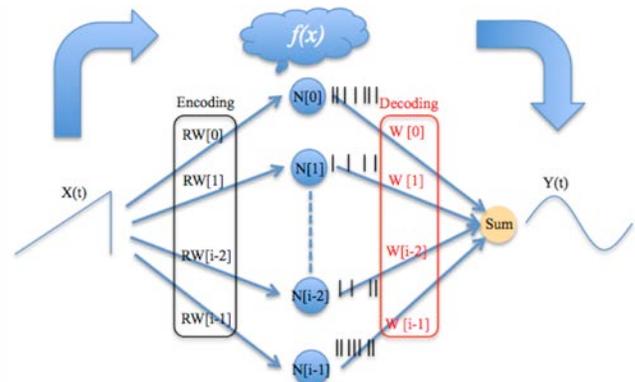

**Figure 1 | A typical NEF network.** The stimulus $X(t)$ is encoded into a large number of nonlinear hidden layer neurons N using randomly initialised connection weights. The output of the system, $Y(t)$, is the linear sum of the weighted spike trains from the hidden neurons.

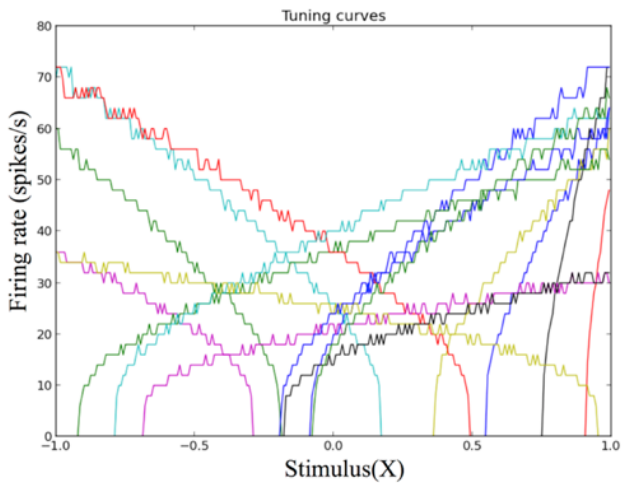

**Figure 2 | Tuning curves maps input stimuli to spike rates.** For clarity, this figure only shows the tuning curve of 16 neurons. Each neuron in the neural layer has a distinct tuning curve.

## 2.2 Algorithm and Theory

*2.2.1 Methodology*

Recognition or classification of handwritten digits is a standard machine learning problem, and in the form of the MNIST database (Lecun et al., 1998) it has become a benchmark problem. Hence, as a proof of concept, we have used the proposed design framework to implement a digit recognition system (Figure 3). Importantly, the same system could be used for other pattern recognition applications. In the MNIST database, the digits are represented as 28 × 28 = 784 pixels, and the training and testing dataset contain 60,000 and 10,000 digits, respectively. The system is trained using the training dataset only and is subsequently validated using the test dataset.

The proposed digit recognition system is a three-layer feed forward neural network, consisting of 784 input layer neurons (pixels), 8192 (8k) hidden layer neurons and ten output layer neurons. The input layer neurons are connected to the hidden layer neurons using randomly weighted all-to-all connections. The hidden layer neurons are also connected to the output-layer neurons using all-to-all connections but with weights calculated using a pseudoinverse operation.

In the digit recognition system, a single input digit (28x28=784 pixels) is mapped onto a layer of input neurons, which we refer to as a vector *Img* with a dimension of 784×1. The *Img* matrix is multiplied by a matrix, *Random_weights*, with a dimension of 8192×784. The resultant vector, referred to as *Vin* with a dimension of 8192×1, is thus given by:

$$Vin = Random\_weights \times Img \qquad (1)$$

Each value in *Vin* is the sum of the randomly weighted pixels, and is the stimulus for the corresponding neuron in the hidden layer. Each neuron of the hidden layer responds to its *Vin* value according to a distinct tuning curve (Figure 2). The output of the hidden layer neurons for each input digit is collected in a matrix referred to as *H* with a dimension of 8192×1. Finally, the response of the output layer neuron is given by:

$$Y = W \times H \qquad (2)$$

where, $W$ is the decoding weight (a matrix with a dimension of 10×8192, ten columns for ten digits: 0-9) and $Y$ (a Boolean matrix with a dimension of 10×1) represents the corresponding value of the input digit. For example, if the input digit represents 2, then, during training, $Y[2]$ will be set to 1 and the other values in $Y$ will be set to 0. Since this is a linear system, the weights can be found by calculating $W = H^+Y$, where $H^+$ is the pseudo-inverse of $H$.

The above description is for one single digit. For training purposes, we used 60000 sample digits and hence the dimensions of *Img, Vin, H* and *Y* will change to 784×60,000, 8192×60,000, 8192×60,000 and 10×60,000, respectively. When we use the digits from the test dataset with 10,000 digits, the dimensions of *Img, Vin, H* and *Y* will change to 784×10,000, 8192×10,000, 8192×10,000 and 10×10,000, respectively. In the testing phase, the predicted output $Y$ will be the product of $W*H$ and will be compared with the expected output to obtain the error rate (the number of unrecognised digits among 10000 test digits). We will address the details of testing in Section 3.

*2.2.2 Modelling*

Our aim is to develop a fast hardware pattern recognition system running in real time, rather than aiming for the lowest test error. Thus, we have adopted a hardware-driven method to implement our system, which will achieve the best trade-off between performance and hardware resources. This method will first consider the hardware

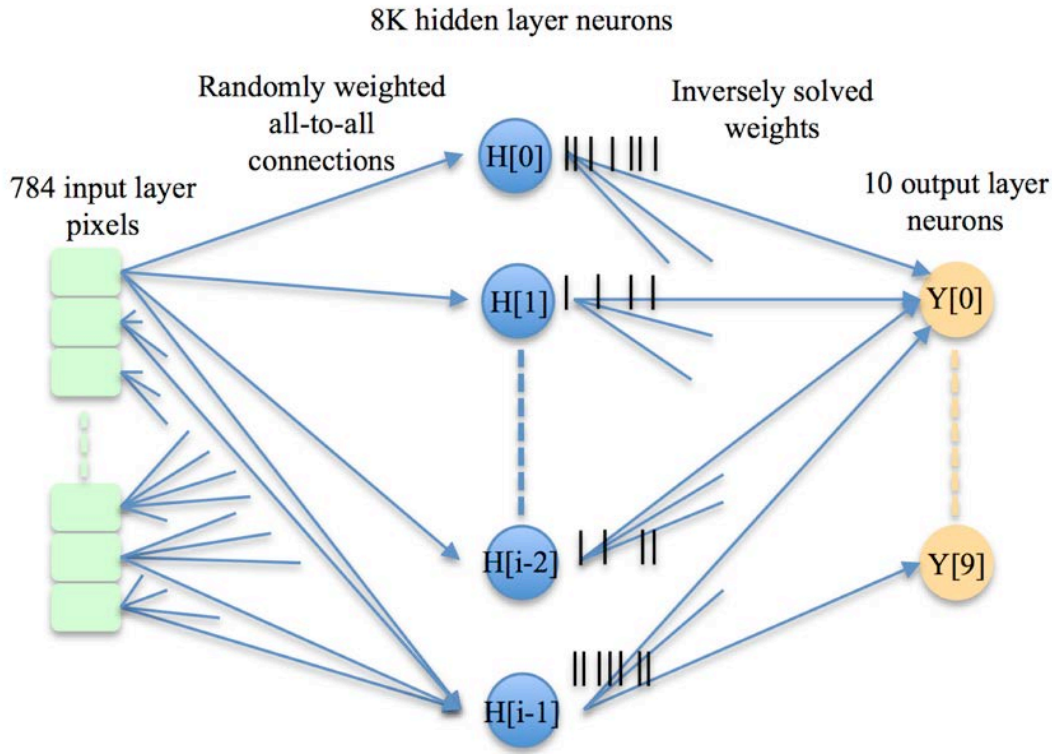

**Figure 3 | System Topology.** The inputs are the pixels; they are connected to a higher-dimensional hidden layer with 8k neurons, using randomly weighted connections. The output layer consists of linear neurons and the output layer weights are solved analytically using the pseudoinverse operation.

constraints, and then all the building blocks will be optimised.

For FPGA implementations, there will be a significant difference in the hardware cost between fixed-point and floating-point implementations, as the latter requires many more digital signal processors (DSPs). More importantly, the floating-point number is represented by 64-bits, which would lead to a huge data storage requirement, which would be a bottleneck for the system. Thus, we have implemented our system using fixed-point numbers.

Before implementing the design in hardware, we have modelled our system in Python, which is a popular software programming language, using the fixed-point representation. This will ensure that the software and the hardware results are the same, and avoid any performance drop or malfunctioning of the system in hardware due to conversion from floating to fixed point numbers. The models presented in the remaining part of this section were all software models unless otherwise specified.

*2.2.3 Input layer*

The input layer will read digits from the MNIST database and map them into the input layer pixels (one by one). This task consists of not only converting the dimension from 28×28 to 784×1 but also converting the grey scale value (an 8-bit number that ranges from 0 to 255) of the pixels to a binary value. The latter is a major difference between our system and existing algorithms (Tapson and van Schaik, 2013) (Lecun et al., 1998). This conversion will reduce the hardware cost significantly with a negligible performance loss, and will be presented in detail in Section 2.3.2. We will compare the performance differences in section 3.1. This conversion is carried out by comparing the grey scale value with 0 - if it is larger than 0, that pixel will be set 1; else it will be set to 0.

To guarantee that the pixels of each digit from the input layer will be nonlinearly projected to the high dimensional hidden layer, for each neuron in the hidden layer, the encoder will first generate a uniformly distributed random weight for each pixel of one input digit and then sum these weighted pixels up for generating the stimulus. For verification of our hardware system, the random weights used in the software and in the hardware models should be the same and produce identical results. In a software model, random weights are generated using special routines, which is difficult to implement on hardware.

One option is to use a look up table (LUT) in the FPGA to store the random weights generated by the software model. The major drawback of this solution is that it requires a significant amount of memory, which scales linearly with number of input neurons and hidden layer neurons. For FPGA implementations, the most efficient way to generate random numbers is to use linear feedback shift registers (LFSRs), as we have previously used to implement a randomly weighed all-to-all connectivity in a spiking neural network (Wang et al., 2014c). Based on that work, we have developed an encoder, which uses LFSRs to perform the nonlinear projection. We have implemented the same LFSR encoder in software to ensure that the random weights are identical in both implementations. We have highly optimised the encoder for hardware implementation, and details of this will be presented in Section 2.3.

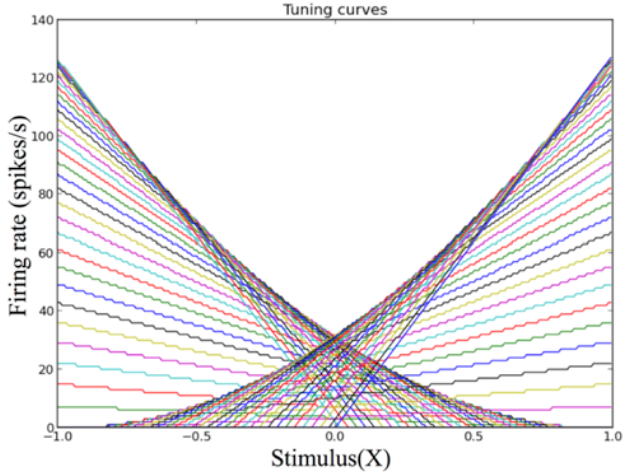

**Figure 4 | The tuning curves of the proposed fixed-point non-spiking neuron.** This figure shows the tuning curve of 64 neurons.

*2.2.4 Rate neuron*

The NEF intrinsically uses spike rates to calculate the weights, and low-pass filters to sum the weighted output spikes to implement the desired function. In contrast, we have implemented our neurons as non-spiking neurons that compute their firing rate directly. If these neurons were to be implemented as leaky-integrate-and-fire neurons on FPGA, as we have done previously (Wang et al., 2014c), their average firing rates would have to be measured for each value of the input stimulus to compute the decoding weights. This method is quite inefficient and inflexible, as we would have to repeat the measurements each time the parameters of the neurons change. Another drawback is that spiking neurons running in real time would not be able to accurately communicate their firing rate in a short time period, e.g., 1ms. This would significantly limit their usage in real time applications. Using non-spiking neurons, their actual firing rate can be communicated immediately after presenting the stimulus to the neurons. This feature is quite important for applications that need real-time control, such as interactive robotic systems.

In a system with non-spiking neurons, the system will not compute correctly if these neurons cannot reproduce the same firing rate as the one used to calculate the decoding weights. In other words, the computed firing rate must be repeatable for a given input value. Based on these requirements, we proposed to compute the firing rate of each neuron using its index in the array together with the stimulus value to produce a 'broken-stick' nonlinearity using the following algorithm:

FOR N_index in (0, N_A-1):

    IF N_index < N_A/2:

        T = Max_Stim - (Stim + 4×N_index)

    ELSE:

        T = Stim + 4×N_index

    F_rate = max(2 × N_index × T / N_A , 0)

END

Here F_rate represents the firing rate of the neuron as a result of the input stimulus, N_index represents the index of the neuron in the neural core, and T is calculated as shown for the different neurons. N_A represents the size of the hidden layer, Max_Stim represents the maximum value of the stimulus and Stim represents the current value of the input stimulus using an integer in the range of [0, Max_Stim) to code for an input range of [-1, 1). Figure 4 shows the tuning curves of a set of N_A = 64 of the proposed fixed-point neurons, using Max_Stim = 255. The transfer function is thus a nonlinear function of the stimulus since the value of F_rate cannot go negative. Our system requires the stimulus to be nonlinearly encoded into the firing rate of the neuron and it is hardware intensive to use digital circuits to implement conventional nonlinear functions such as *tanh*. Instead, this piecewise linear function can be easily implemented using a single 9-bit fixed-point multiplier. We will present its implementation in detail in section 2.3.3.

*2.2.5 Hidden layer*

We refer to the set of 64 neurons as a neural core, which will be used as the standard building block for our digit recognition system. Multiple neural cores can easily be combined to build real-time large-scale neural networks using our design framework. Furthermore, the development cycle of large-scale neural networks will be significantly shortened as there is no requirement for measurement of the firing rate anymore, since each neural core has the same set of known tuning curves.

The hidden layer was implemented with 128 identical neural cores, for a total of 8192 (8k) neurons and 8192×(784+10) ≈ 6.5M synaptic connections. This hidden layer size has achieved the best trade-off between performance and memory usage and we will compare the performance differences in Section 3.2. Given an input image, the encoder will generate, via the random weight projection, a different *Vin* for each neuron in each core, even if each core contains identical neurons. In other words, even though neuron[0] in neural core[0] and neuron[0] in neural core[1] have the same tuning curve as a function of *Vin*, the are highly likely to get different *Vin* so that their firing rates will be different too.

*2.2.6 Regression*

The decoding weights are obtained by calculating $W = H^+Y$, where $H^+$ is the pseudoinverse of *H*. However, the pseudo-inverse of the matrix *H* of size 60000 × 8192 requires a huge amount of memory and computational time. We have previously developed an online pseudoinverse update method (OPIUM) (Tapson and van Schaik, 2013), which is an incremental method to compute the pseudoinverse solution to the regression problem, which requires significantly less memory. Hence, we use this method here to compute the decoding weights. We chose to use a 6-bit resolution for the decoding weights, to obtain the best trade-off between performance and memory usage. We will address this in details in section 3.1.

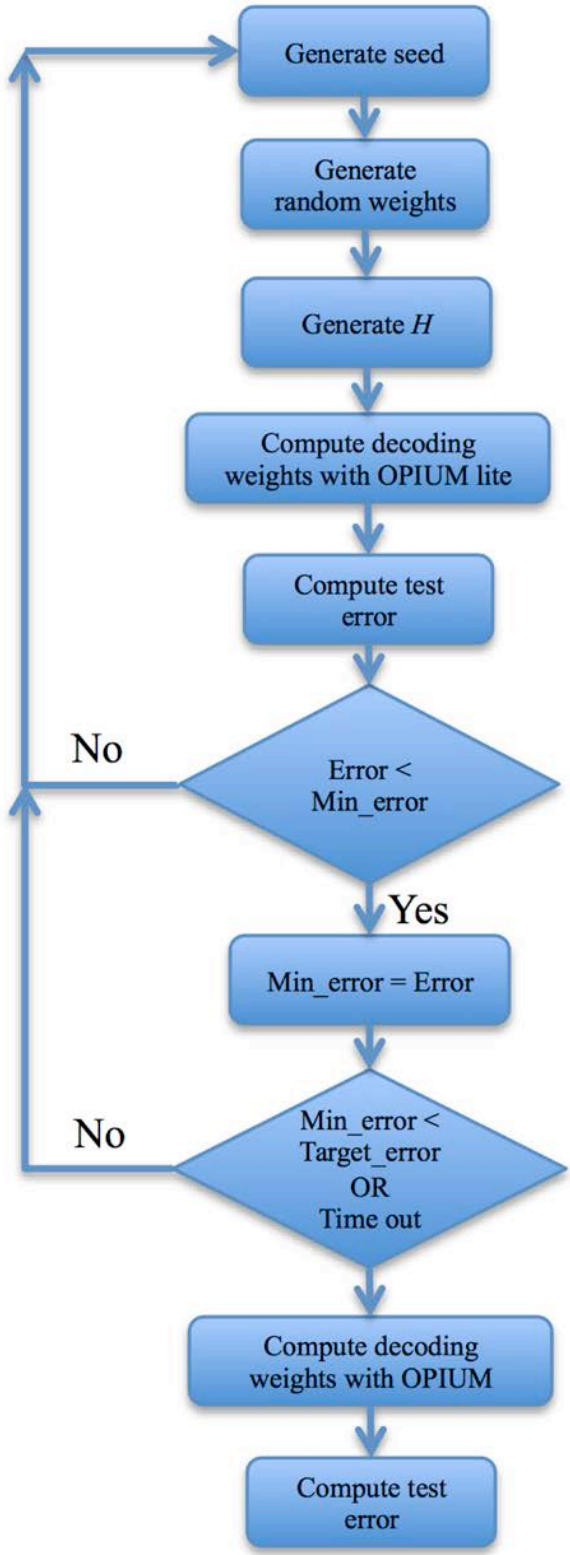

**Figure 5 | The flow of the proposed regression method.**

The pseudoinverse method only gives the best solution with the lowest square root error for any given $H$ matrix, i.e., any given set of random weights; it does not necessarily achieve the lowest test error for the MNIST data set. So we adopted a regression method to find the best seed, which will be used by the encoder to generate random weights, and will in turn change the $H$ matrix. In this way, we can obtain the lowest possible test error in our system. Figure 5 shows the flow of this regression method. It uses a simplified version of OPIUM, called OPIUM lite (Tapson and van Schaik, 2013), which is a fast online method for calculating an approximation to the pseudoinverse. It is significantly quicker than the full-scale OPIUM, but will find output weights resulting in a slightly worse test error. OPIUM lite is used with different random seeds, i.e., for different random weight vectors, until a seed is found with a target error below a desired threshold. After that, the full scale OPIUM is used to compute the decoding weights with that seed. As there is no guarantee that OPIUM lite will be able to achieve a target error below the desired threshold, a time-out mechanism is introduced. In our system, this time-out will be activated when the regression has run for 1000 seeds. If a time-out happens we simply use the seed that has so far resulted in the lowest error and then use the full scale OPIUM to compute the decoding weights.

## 2.3 Hardware implementation

### 2.3.1 Topology

To efficiently implement the system on an FPGA, we use a time-multiplexing approach (Cassidy et al., 2011; Wang et al., 2013, 2014d, 2014c, 2014b, 2015; Thakur et al., 2014), which leverages the high-speed digital circuit. State-of-the-art FPGAs can easily run at a clock speed of 266MHz (clock period 3.75ns). Thus, we can exploit time-multiplexing approach to simulate $2^{18}$ neurons (256k, powers of two are preferable as they optimise memory use for storage) in ~1 millisecond by only implementing one physical neuron on an FPGA. We refer to these neurons as time-multiplexed (TM) neurons. This means that on every clock cycle, a TM neuron will be processed. Each TM neuron is updated every 256k/266MHz ≈ 943 μs while a sub-millisecond resolution is generally acceptable for neural simulations.

The time-multiplexing approach is however constrained by its data storage requirement. The on-chip SRAM is limited in size (usually only tens of MBs). Due to bandwidth constraints it is difficult to use off-chip memory with the time-multiplexing approach, as new values need to be available from memory every clock cycle to provide real-time simulation. Furthermore, the architecture of the system will be more complex when using off-chip memory because it needs a dedicated memory controller. Nevertheless, using off-chip memory promises the ability to implement much larger networks and we will investigate this option for future designs. However, we chose to use on-chip memory for the current work to keep the architecture simple.

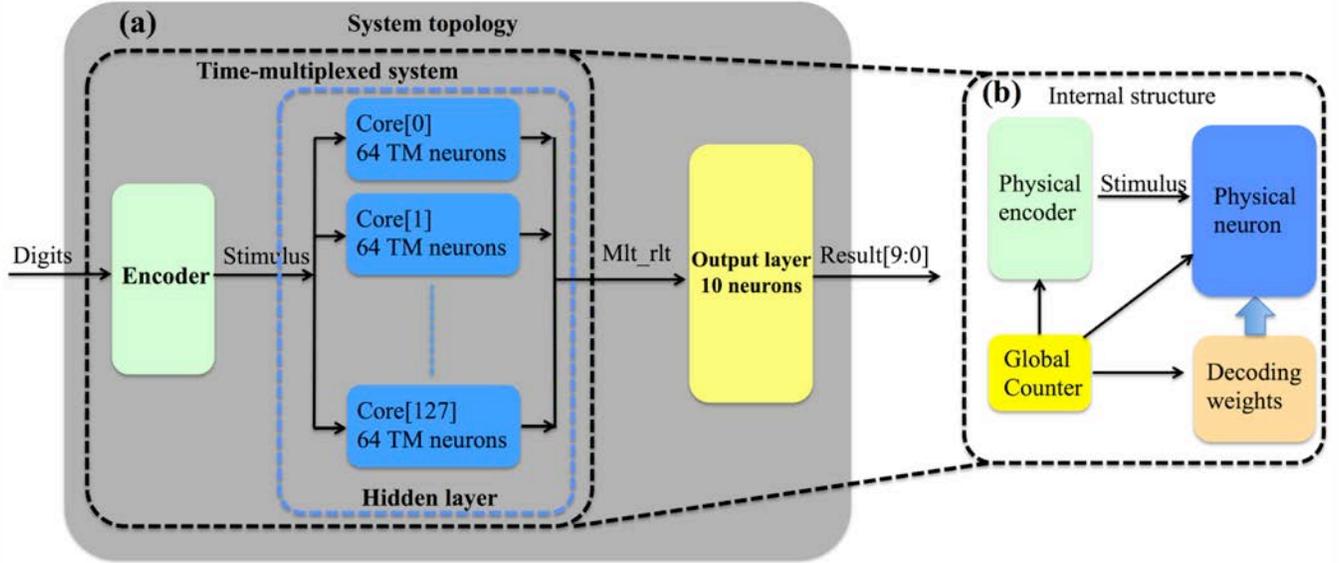

**Figure 6 | FPGA implementation of the proposed system.** (a) The system topology;(b) The internal structure of the time-multiplexed system.

Figure 6 shows the topology of the FPGA implementation of the system, which consists of an input layer (the encoder), a hidden layer with 128 neural cores and an output layer with 10 neurons. The encoder and the hidden layer are both implemented with the time-multiplexing approach and Figure 6b shows their internal structure. It consists of a physical encoder, a physical neuron, a global counter and a weight buffer. The global counter processes the time-multiplexed (TM) encoders and neurons sequentially. The decoding weights of the physical neuron are stored in the weight buffer. For simplicity, let us assume that each TM encoder and TM neuron are processed in only one clock cycle. This means that in every clock cycle, a TM encoder will generate the stimulus for an input digit, and the corresponding TM neuron will generate a firing rate with that stimulus and then multiply it with the decoding weights (ten numbers for ten digits obtained by using the OPIUM). The input digit will not change and will remain static until all the TM neurons finish their processing. The output of every TM neuron will be ten weighted firing rates, each of which will be accumulated by its corresponding output neuron. Using a pipelined architecture, the result from calculating one time step for a TM encoder and neuron only has to be available just before the turn of that TM encoder and TM neuron comes around again. The above description assumes that it only takes one clock cycle to process one TM encoder and TM neuron, while this timing requirement is quite difficult to meet in a practical design. We will address this issue in detail in next section.

*2.3.2 Physical encoder*

The encoder will generate a uniformly distributed random weight for each pixel of the input digit, and then sum these weighted pixels to generate the stimulus for each neuron in the hidden layer. We have pre-processed the input digit by converting grey-scale value of each pixel to a binary value. This saves significant hardware resources in the FPGA, since otherwise we would need 784 multipliers to compute the multiplication between all pixels and their corresponding random weights. Each binary pixel is used to control a 2-input multiplexer, one is connected to its corresponding random weight and the other is tied down to zero. If the value of a pixel is high, that corresponding random weight will be accumulated for the generation of stimulus for a hidden layer neuron.

The major challenge in implementing the encoder in hardware using the time-multiplexing approach is to meet the timing requirement. We need to sum all the 784 weighted pixels in 3.75 ns, since each TM neuron needs to be processed in one clock cycle. Moreover, this operation will require 784 adders, which will cost a significant amount of hardware resources. The introduction of pipelines will mitigate the critical timing requirement, but will need even more adders. As a compromise we chose to process each TM encoder and TM neuron in a time slot of four clock cycles. So the encoder will perform this sum operation in four cycles, each of which will sum 784/4=196 weighted pixels. This modification not only mitigates the critical timing requirement, but also reduces the number of adders that are needed. The price paid is that the time-multiplexing rate has to be divided by four. Hence, we can only time-multiplex 64k neurons rather than 256k neurons.

Figure 7 shows the structure of the physical encoder, which consists of an input buffer, a global counter, 49 random weight (RW) generators (each implemented with an 20-bit LFSR), 196 2-input multiplexers and a sum up module. When an input digit arrives, it is stored in the input buffer. In each time slot, the global counter sends that stored digits to multiplexers for generating the weighted pixels. The lowest 196 bits are sent in the first clock cycle (of that time slot) and then the higher 196 bits in the next clock

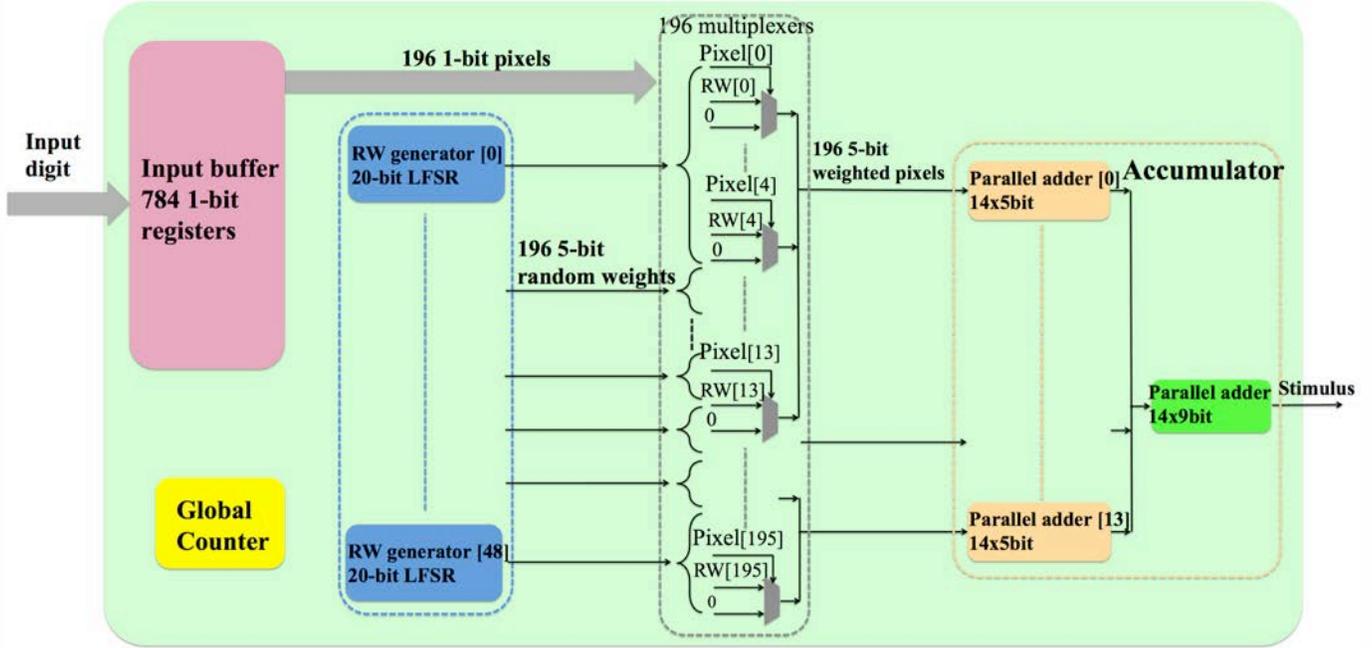

**Figure 7. The structure of the physical encoder**

cycle, one by one, and highest 196 bits in the fourth clock cycle.

Each RW generator generates a 20-bit random number, which is divided into four 5-bit random signed numbers. Hence, 49 RW generators will provide totally 49x4 = 196 5-bit random weights, each is sent to its corresponding multiplexer. All these LFSRs will reload their own initial seed (obtained using the pseudoinverse method) on the arrival of an input digit. After that, it keeps generating random numbers until a new input digit arrives. In this way, we can guarantee that the encoder will generate the exact same set of random weights (for each incoming digit) with any given seed. This "on the fly" generation scheme reduces the usage of the memory significantly, as there is no requirement for storing the random weights anymore – only the seeds need to be stored.

The accumulator module sums the 784 weighted pixels (in four clock cycles) for generating the stimulus for that TM neuron. A naive implementation would need a 196-input 5-bit parallel adder and create a large delay (~20 ns). To mitigate this critical timing requirement, we use a 2-stage pipeline, which consists of fourteen 14-input 5-bit parallel adders and one 14-input 9-bit parallel adder. Since it is a pipelined design, the stimulus (for each TM neuron) is still being generated every time slot (with a latency of two clock cycles).

*2.3.3 Physical neuron*

The rate neuron achieves a significant reduction in memory usage, since it computes its firing rate with its index, the input stimulus and fixed parameters, none of which need memory access. Memory access is only needed to read the decoding weights. In our previous work (Wang et al., 2014a), the physical neuron has already been implemented with a single 9-bit multiplier, which computes the F_rate and multiplies it with one and only one decoding weight. In the digit recognition system implemented here, the neuron needs to multiply F_rate with ten decoding weights (for ten digits: 0-9). A naïve implementation would instantiate ten identical neurons, each with one decoding weight (for each output neuron), and would cost 10 multipliers. The whole operation would require 11 multiplications. Since the time slot consists of four clock cycles, we can distribute these 11 multiplications to these four clock cycles so that only 11/4=3 multipliers will be needed. Based on this strategy, the neuron has been efficiently implemented with three identical 9-bit multipliers as shown in Figure 8. The number of the implementable multipliers is usually one of the bottlenecks

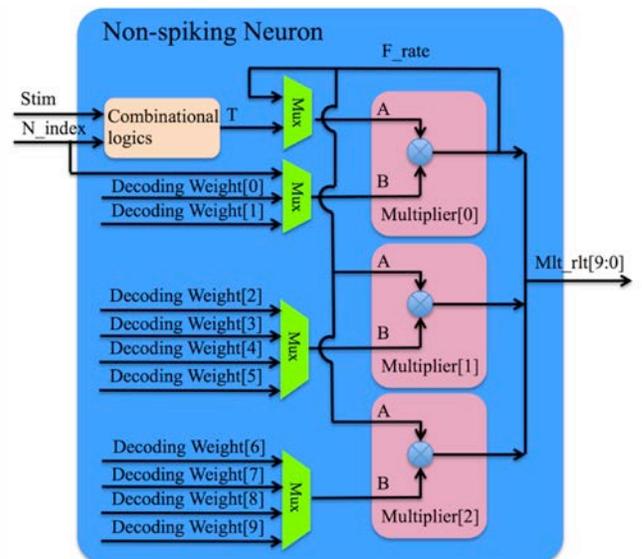

**Figure 8. The structure of the physical neuron**

TABLE I

Device utilisation Altera Cyclone 5CGXFC5C6F27C7

| Adaptive Logic Modules (ALMs) | RAMs | DSPs |
|---|---|---|
| 2162/29080 | 480k/4.5M | 3/450 |

of large-scale FPGA/ASIC design.

The multiplier's inputs A and B are 9 bits wide and the output result is 18 bits wide. All of the three multipliers will need four clock cycles to process the algorithm. For multiplier [0], the first cycle computes the F_rate, which is represented by a 7-bit number, by multiplying N_index and T; the second cycle latches F_rate at input A of the multiplier; the third and fourth cycle multiplies F_rate with the decoding weight [0] and [1], respectively. For multiplier [1], the first, second, third and fourth cycle multiplies F_rate with the decoding weight [2],[3],[4] and [5] respectively. For multiplier [2], the first, second, third and fourth cycle multiplies F_rate with the decoding weight [6],[7],[8] and [9] respectively. Again, since it is a pipelined design, the output of each TM neuron is updated only once in its time slot (with a latency of four clock cycles).

*2.3.4 Output layer*

The output layer consists of ten neurons (Figure 6) that will linearly sum the results of all the 8k TM neurons. Since it is a time-multiplexed system, this sum is just an accumulation of the outputs of the TM neurons of each time slot and the computational cost can be reduced in magnitudes. Hence, the implementation of each output neuron will only need a register and an adder. When all the 8k neurons have all been processed, the index of the output neuron with the maximum value will be sent out as the result, which indicates the most likely input digit. After that, the values of the ten output neurons are cleared.

*2.3.5 Utilisation*

The system was developed using the standard ASIC design flow, and can thus be easily implemented with state-of-the-art manufacturing technologies, should an integrated circuit implementation be desired. A bottom-up design flow was adopted, in which we designed and verified each module separately. Once the module level verification was complete, all the modules were integrated together for top-level verification. We have successfully implemented 128 proposed neural cores, yielding 8k neurons, on an Altera Cyclone V FPGA (on a Terasic Cyclone GX starter kit). The design uses less than 6% of the hardware resources (with the exception of the RAMs, Table I). Note that this utilisation table includes the circuits that carry out other tasks such as the JTAG interface.

## 3. Results

The results presented here will focus on how different design choices will affect the performance of the proposed system as our goal is to develop a hardware system running in real time, rather than exploiting an algorithm that is as accurate as possible. The performance results were obtained using the full test set of 10,000 handwritten digits after training on the full 60,000 digit training set, unless otherwise specified. The results presented in Section 3.1-3.2 were all obtained using the software (Python) models. The results presented in section 3.3 were obtained from the hardware implementation.

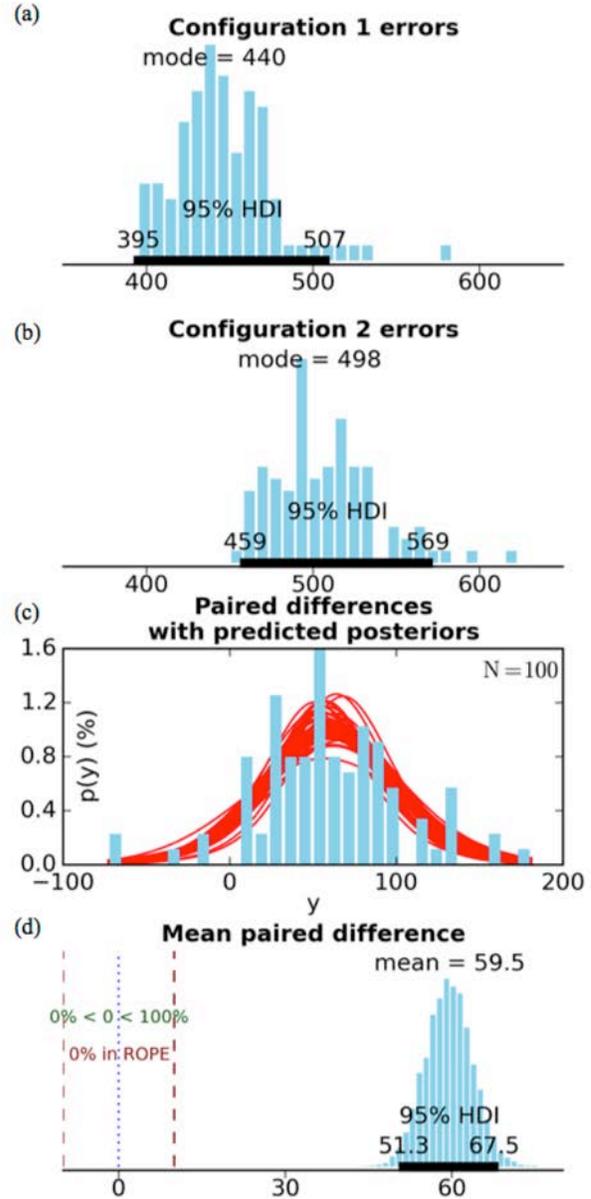

**Figure 9.** (a) and (b) The histogram of the error rate for configuration 1 and configuration 2; (c) the normalised histogram of the difference between the paired errors (blue) and sample T distributions modelling the data (red); (d) the distribution of the estimated mean of the difference data.

*3.1 Comparison across different configurations*

Compared to our previous work (Tapson and van Schaik, 2013), we have made three major modifications: the grey-scale pixel in the input images were replaced by black & white (binary) pixels; *tanh* neurons in the hidden layer were replaced by rate neurons; and 64-bit floating-point numbers for the decoding weights were replaced by 6-bit fixed-point numbers. We investigated the effects of these modifications using four configurations: configuration 1 was the configuration used in our previous work (Tapson and van Schaik, 2013); configuration 2 used black and white images; configuration 3 used black and white images and rate neurons instead of *tanh* neurons; and configuration 4 had all three modifications. The hidden layer consisted of 8k neurons in all four configurations.

For each configuration, 100 test runs were conducted, each with a different random seed. The same set of 100 seeds was used for all four configurations, so that the encoder will generate the same random weights. Since the goal of this exercise was simply to investigate the impact of the three modifications on performance, rather than to find the best possible performance, we only used the first five steps of the regression method, i.e., we only used OPIUM lite to calculate the decoding weights and the test error. This significantly reduces the simulation time needed for these tests while still providing a fair comparison between the four configurations.

We first investigated the effect of using the binary values in the input layer. We compared the performance result between the one using the grey-scale values and binary values (see Figure 9). The top two panels show a histogram of the number of errors out of 10,000 test patterns. Given the skewed nature of the two error distributions, rather than simply reporting p-values to indicate the statistical significance of this difference, we have chosen to display the full distribution here. Because the same set 100 random weight vectors was used for each configuration, we can determine a paired difference between the two configurations, shown as a histogram in Figure 9c. We then modelled the distribution of the difference of errors using a non-central T distribution, which is optimal for modelling distributions that are approximately Gaussian but contain outliers. We followed the Bayesian estimation method according to Kruschke (Kruschke, 2012) using Markov Chain Monte Carlo simulation. We simulated the Markov Chain for 110,000 steps and discarded the first 10,000 steps as a burn in period. Figure 9d shows the distribution of the 100,000 mean values for the T distribution modelling the data, and the red curves in Figure 9c show 50 examples of the T distribution with parameters (mean, standard deviation, and a normality parameter – see (Kruschke, 2012)) taken at random from the Markov Chain.

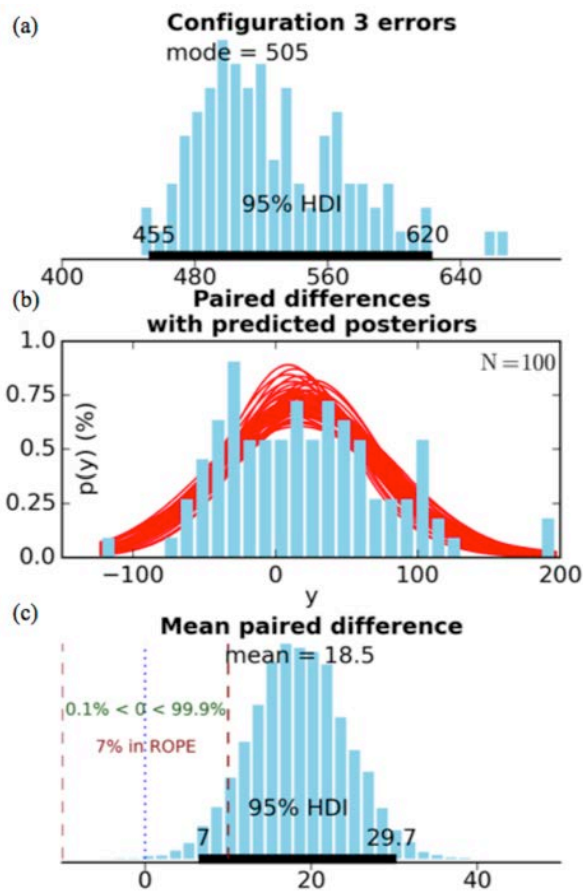

**Figure 10.** (a) The histogram of the error rate for configuration 3; (b) the normalised histogram of the difference between the paired errors (blue) and sample T distributions modelling the data (red); (c) the distribution of the estimated mean of the difference data.

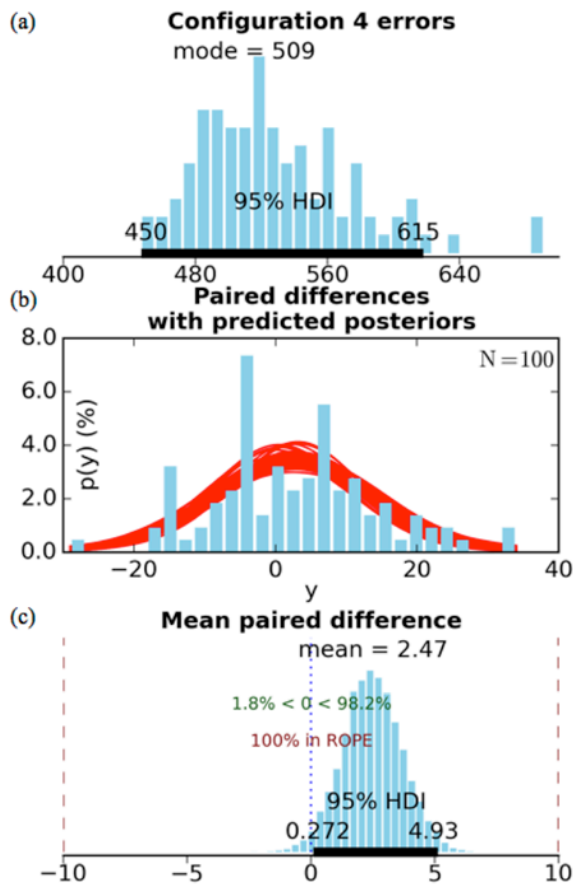

**Figure 11.** (a) The histogram of the error rate for configuration 4; (b) the normalised histogram of the difference between the paired errors (blue) and sample T distributions modelling the data (red); (c) the distribution of the estimated mean of the difference data.

From the distribution of the mean value for the difference data (Figure 9d), we can see that configuration 2 results in 59.5 more errors on average. If we define a difference of 10 or fewer errors as a region of practical equivalence (ROPE), or, in other words, we consider as insignificant a change of 10 or fewer errors out of 10,000 tests, i.e., a change of less than 0.1%, we note that the 95% highest density interval (HDI) of the distribution of the mean of the difference of errors is outside the ROPE, and therefore we conclude that changing the input images from grey scale to binary values results in a small but significant increase in error of around 0.6%.

Next, we investigated the effect of using the rate neurons in the hidden layer. The distribution of errors for this configuration (configuration 3) is shown in Figure 10a. This should be compared with configuration 2 (Figure 9b) and their paired difference is shown in Figure 10b. Figure 10c shows the distribution of the mean of the difference in errors between configuration 3 and configuration 2. It shows that changing from *tanh* neurons to rate neurons increases the number of errors by approximately 18.5. However, this difference is not strongly significant, as the 95% HDI is not entirely outside the ROPE, indicating that a difference within the region of practical equivalence is amongst the possible mean values. Finally, we investigated the effect of using limited-resolution decoding weights. Figure 11a shows the distribution of errors for this configuration and the difference between configuration 3 and configuration 4 is close to zero (Figure 11b). In fact the distribution of the mean of the error difference is entirely within the ROPE, indicating that somewhat surprisingly there is no significant loss in performance when using 6-bit fixed-point output weights instead of floating point weights.

The performance drop between configuration 1 and 4 was merely 0.8%. We can therefore conclude that, in this digit recognition system, the modifications that we made achieved significant reductions in terms of hardware cost with a minimal drop in performance.

*3.2 Size of the hidden layer*

In this scenario, we used configuration 4 from the previous section and changed the hidden layer size in the range from 1k to 16k neurons. For each size, ten test runs (each with a different random seed) were conducted. Again, to reduce the testing time, we used OPIUM lite to calculate the decoding weights and then calculate the test error.

The median error over 10 runs (Figure 12) for the hidden layer with 1k, 2k, 4k, 8k, 12k and 16k neurons was 14.5%, 10.4%, 6.96%, 5.01%, 4.47% and 4.33% respectively. It is clear that the error decreases with the number of hidden layer neurons, although with a diminishing return. Since the system used the time-multiplexing approach and rate neurons, the hardware cost of a single TM neuron is almost negligible. The memory required by the decoding weights is linearly proportional to size of the hidden layer and is thus the bottleneck of the system. To achieve a good balance between the desired accuracy and memory, we chose to implement the hidden layer with 8k rather than 16k neurons.

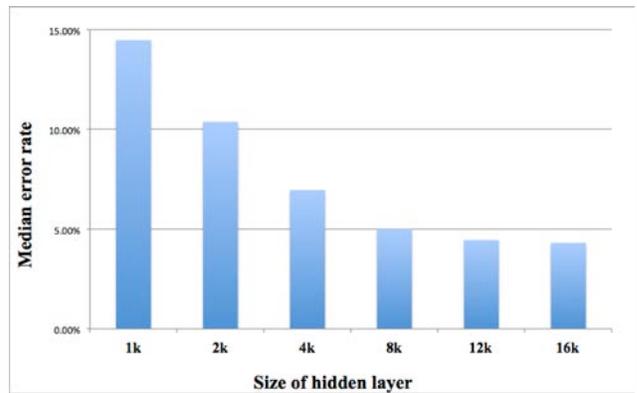

**Figure 12. Error rates as a function of the number of neurons in the hidden layer.**

*3.2 System performance*

To explore the best performance that the proposed system can achieve, 1000 runs were carried out using the full regression method (Figure 5) with different random seeds. The lowest error achieved with lite and full version of OPIUM is 4.52% and 3.45%, respectively. After that, the decoding weights (obtained with full version of OPIUM) were loaded into the FPGA board for real time digit recognition. The pixels of input digits were converted to binary values in software and a Python-based front-end client software sent the selected test digit to the FPGA via JTAG interface. Since the system runs at 266MHz and the hidden layer contains 8k neurons, each of which has a time slot of four clock cycles, the processing time for one input digit will be 8k×4/266MHz ≈ 120 μs, yielding 1s/120μs ≈ 8k digit recognitions per second. Due to the fact that our system only used 8k out of 64k neurons in one single TM neuron layer, the maximum number of the digit recognitions that can be processed by one TM neuron layer is ~64k per second. The system used less than 6% of the hardware resources (with the exception of the RAMs), multiple TM neuron layers can be instantiated to run in parallel. It is practical to scale this system to process millions of digit recognitions in one second. We will address this in details in section 4.2.

## 4. Discussion
### 4.1 Comparison with other solutions

The work reported here constitutes the basis for building real-time, large-scale, general purpose hardware pattern recognition systems using the NEF, hence we are mainly interested in the trade-off between the scale, the performance and the hardware cost. We will concentrate on comparing our work with the solutions that were developed for similar goals, rather than the solutions that are extremely optimised for achieving the lowest error rate of MNIST although they cannot be efficiently implemented on hardware.

The IBM TrueNorth system is a general-purpose system for building large-scale neural networks running in real time (Merolla et al., 2014). When it was programmed for digit recognition, it achieved a result of 8.06% error rate in the 10000 test set of the MNIST with 13 cores, each of which consisted of TM 256 spiking neurons and needs ~96k bits memories (Esser et al., 2013). Hence, our system achieved a much lower error rate while with significantly fewer hardware resources, especially the memories (Table II). Regarding the processing speed, their system needs 20 time steps (each one is 1 ms) to process one digit, whereas our system needs only 120 µs (approximately 167 times speedup). Moreover, while their system consists of a feature extractor that clusters and extracts features from data, our system is feature-less, hence can be easily configured for different input data without feature extractions. The TrueNorth system however has much more applications besides pattern recognition task, as compared to our system.

The Minitaur, which is an event–based neural network accelerator, achieved an error rate of 8% on a deep spiking network with 1785 neurons (Neil and Liu, 2014). Since the scheme it used is a variant of the time-multiplexing approach, which only needs very few neurons to be physically implemented, the cost of one single neuron is also negligible and the bottleneck again is the memory. Each of the neuron used by the Minitaur needs 73 bits memories and the connection weight needs 16 bit memories. Our neuron needs 60 bit memories for the decoding weights. The processing time of the Minitaur for one digit is 0.152s (table II), which is approximately 1300 times slower than our system.

### 4.2 Future work

Since the larger the scale is, the more pattern recognitions can be carried out, our future work will focus on scaling up the network that we have presented here. It is a scalable design as it is a fully digital implementation. The number of TM hidden neurons implemented by a single physical neuron will increase linearly with the amount of available memory, as long as the multiplexing scale keeps the time resolution within the biological time scale. The number of physical neurons will increase linearly with the number of available ALMs.

In the following calculation, we will use the digits recognition system as a metric and different applications will require different amounts of hardware resources while still using the same topology. We can calculate the theoretical maximum network size on a state-of-the-art FPGA board, such as the Terasic DE5 board containing an Altera Stratix V (5SGXEA7N2F45C2) FPGA with ~230k ALMs, two DDR3 SDRAMs and four QDRII+ SRAMs. One single TM hidden layer requires ~1600 ALMs, which

TABLE II

Comparison with other solutions

|  | Error | Computation time | Resources |
|---|---|---|---|
| Minitaur | 8% | 0.152 s | 155k bits |
| TrueNorth | 8.06% | 20 ms | 1.248M bits |
| This work | 3.45% | 120 µs | 480k bits |

is mainly used by the encoders. Hence, the maximum number of the physical hidden neurons that can be implemented is 230k/1600 ≈ 143. The memory requirement of one single TM hidden neuron layer is 64k×60bits = 3840k bits. The on-chip SRAM, which is 52M bits, can be used to implement up to 13 TM hidden neuron layers. To further scale up the system, we need to use external memories. The bandwidth requirement is indeed a bottleneck for the time-multiplexing approach, as new values need to be available from memory every four clock cycles.

The maximum theoretical bandwidth of one DDR3 SDRAM memory and one QDRII+ SRAM memory on the DE5 board is 512 bits and 72 bits @266MHz, respectively. The DDR3 memory, in general, can only achieve an efficiency of 70% (of the theoretical bandwidth) as it will need flow control, which takes into consideration the bus turn around time, refresh cycles, and so on. The maximum number of neuron arrays is ((512bits × 2 × 70% + 72bits×4)×4)/60bits ≈ 67. Adding the ones using the on-chip SRAM, the theoretical maximum number of neuron layers is 80, yielding 64k×80 = 5.12M neurons. As the maximum number of the digit recognitions that can be processed by one TM neuron layer is ~64k per second, the maximum number of the digit recognitions that can be processed by the system with 80 parallel layers is therefore 5.12M per second.

The programmability of the FPGA, especially the decoding weights, makes the integration of the system with the desired pattern recognition applications seamless. However, the advantages of running large-scale networks in real-time are strongly reduced if such neural networks take a long time to compute the decoding weights. Hence, another major improvement is to speed up this computationally extensive task. One promising solution is to implement the OPIUM on FPGA, since this algorithm is an adaption procedure without the requirement of hundreds of Gigabyte RAMs and is quite friendly for hardware implementation. Running OPIUM in real time makes it possible to upgrade the system to be a true turnkey solution for pattern recognition in real world. In addition, since the proposed system does not need feature extraction, it could be used for any other pattern recognition tasks such as speaker recognition, natural language processing and so on.

# 5. Acknowledgment

This work has been supported by the Australian Research Council Grant DP140103001. The support by the Altera university program is gratefully acknowledged. This work was inspired by the Capo Caccia Cognitive Neuromorphic Engineering Workshop 2013, 2014 and Telluride Neuromorphic workshop 2013.